\documentclass{article}

\usepackage{microtype}
\usepackage{graphicx}
\usepackage{subfigure}
\usepackage{booktabs} % for professional tables

\usepackage{hyperref}

\usepackage[accepted]{icml2024}

\usepackage{amsmath}
\usepackage{amssymb}
\usepackage{mathtools}
\usepackage{amsthm}
\usepackage{multirow}
\usepackage{makecell}
\usepackage[capitalize,noabbrev]{cleveref}
\usepackage{nccmath}

\theoremstyle{plain}

\theoremstyle{definition}

\theoremstyle{remark}

\newcommand{\toxicprobesymbol}{$W_{\text{Toxic}}$}
\newcommand{\toxicmlpsymbol}{$\text{MLP.}\mathbf{v}_{\text{Toxic}}$}
\newcommand{\toxicmlpkeysymbol}{$\text{MLP.}\mathbf{k}_{\text{Toxic}}$}
\newcommand{\toxicsvdsymbol}{$\text{SVD.U}_{\text{Toxic}}$}

\newcommand{\dpo}{$\text{GPT2}_{\text{DPO}}$}

\newcommand{\TODO}[1]{{\color{red}[\textbf{TODO}]}}
\icmltitlerunning{A Mechanistic Understanding of Alignment Algorithms}

\begin{document}

\twocolumn[
\icmltitle{A Mechanistic Understanding of Alignment Algorithms:\texorpdfstring{\\}{}
A Case Study on DPO and Toxicity}



\icmlsetsymbol{equal}{*}

\begin{icmlauthorlist}
\icmlauthor{Andrew Lee}{umich}
\icmlauthor{Xiaoyan Bai}{umich}
\icmlauthor{Itamar Pres}{umich}
\icmlauthor{Martin Wattenberg}{harvard}
\icmlauthor{Jonathan K. Kummerfeld}{sydney}
\icmlauthor{Rada Mihalcea}{umich}
\end{icmlauthorlist}

\icmlaffiliation{umich}{University of Michigan, Ann Arbor, U.S.A.}
\icmlaffiliation{harvard}{Harvard University, Cambridge, Massachusetts}
\icmlaffiliation{sydney}{University of Sydney, Sydney, Australia}

\icmlcorrespondingauthor{Andrew Lee}{ajyl@umich.edu}

\icmlkeywords{Machine Learning, ICML, mechanistic interpretability, alignment}

\vskip 0.3in
]



\printAffiliations{}
%\printAffiliationsAndNotice  % leave blank if no need to mention equal contribution

\begin{abstract}

While alignment algorithms are now commonly used to tune pre-trained language models towards a user's preferences, we lack explanations for the underlying mechanisms in which models become ``aligned'', thus making it difficult to explain phenomena like jailbreaks.
In this work we study a popular algorithm, direct preference optimization (DPO), and the mechanisms by which it reduces toxicity.
Namely, we first study how toxicity is represented and elicited in a pre-trained language model, GPT2-medium.
We then apply DPO with a carefully crafted pairwise dataset to reduce toxicity.
We  examine how the resulting model averts toxic outputs, and find that capabilities learned from pre-training are not removed, but rather bypassed.
We use this insight to demonstrate a simple method to un-align the model, reverting it back to its toxic behavior.

\end{abstract}

\section{Introduction}
\label{sec:intro}
Large language models learn surprising capabilities from pre-training on large datasets \citep{gpt2FewShotLearners, chowdhery2023palm, touvron2023llama}.
While these capabilities lead to impressive achievements, they also include unwanted behaviors that can be found in large-scale web data, such as toxicity and bias \citep{sheng-etal-2019-woman, gehman-etal-2020-realtoxicityprompts}.
As a result, researchers have developed alignment algorithms to reduce undesirable behaviors, which often use reinforcement learning with human preferences (RLHF).
For instance, proximal policy optimization (PPO, \citealt{schulman2017proximal}) fits a reward model on human preference data, which is then used to fine-tune a language model, while direct preference optimization (DPO, \citealt{rafailov2023direct}) by-passes the reward model and derives reward signals directly from pairwise preference data.

While such algorithms can suppress undesirable behavior, our understanding of the mechanisms by which the undesirable behavior is suppressed is limited.
Furthermore, researchers have demonstrated that such alignments can be surprisingly easily undone \citep{wallace-etal-2019-universal, zou2023universal, wei2023jailbroken, carlini2023are}.
While prior work hypothesize why jailbreaks are possible through empirical studies~\citep{wei2023jailbroken}, in this work we provide a mechanistic explanation for such phenomena.

Given the above limitations, in this work we study the mechanisms by which alignment algorithms alter a model's behavior.
Researchers have demonstrated that a deep enough understanding of a model's inner representations allows us to interpret how it makes decisions.
For instance, various concepts such as world models, truthfulness, or even task-specific features have highly interpretable and controllable representations \citep{li2023inference, todd2023function, nanda2023emergent}.
Motivated by such findings, we study how the representation space of language models change by comparing it before and after an alignment algorithm is applied.
Our work relates to that of \citet{jain2023mechanistically}, which studies how the capabilities of a language model changes after fine-tuning on synthetic tasks.
Unlike this previous work, we study the change in mechanisms from a RLHF algorithm on a natural language setting.

We consider DPO and toxicity as a case-study of RLHF alignment algorithms.
We first study how toxicity is represented and elicited in GPT2-medium (henceforth GPT2).
We then apply DPO using a carefully crafted pairwise dataset that consists of toxic and nontoxic samples.
Lastly, we study the mechanisms by which toxicity is no longer generated after DPO, and how those mechanisms can fail.

Our work is organized as follows: in Section~\ref{sec:preliminaries} we provide the necessary preliminaries relevant to our work.
In Section~\ref{sec:gpt2_toxic_representation}, we demonstrate how toxicity is represented and elicited in GPT2.
We find multiple vectors in multilayer perceptron (MLP) blocks that promote toxicity.
We apply singular value decomposition (SVD) to these toxic vectors to find vectors that represent specific dimensions of toxicity in the model.
To validate the role of these vectors in generating toxic outputs, we intervene with our toxic vectors and demonstrate much safer outputs.

In Section~\ref{sec:dpo}, we explain our procedure to apply DPO on our language models to reduce toxicity, using a carefully crafted pairwise toxicity dataset, produced by using PPLM~\citep{dathathri2019plug} to generate paired toxic and non-toxic samples.

In Section~\ref{sec:post_dpo}, we demonstrate how toxicity is no longer elicited after DPO.
Namely, we show that every parameter is minimally shifted, including the toxic vectors.
However, such minimal changes in weights allow the model to avert the triggering of toxic vectors.
Put differently, DPO \emph{does not remove} the capability of generating toxic outputs, but learns an ``offset'', distributed amongst its layers, to ``bypass'' the regions that elicit toxicity.
Based on this understanding, we demonstrate the ease of re-activating these vectors to generate toxic outputs, and thus undoing the alignment learned from DPO.
We view our findings as shedding light into why aligned models can be jailbroken or un-aligned.

\section{Preliminaries}
\label{sec:preliminaries}

In this section we provide background and notations, much of which is borrowed from \citet{geva-etal-2022-transformer}.

\paragraph{Transformers, MLPs.}
Transformer-based language models typically consists of embedding and unembedding layers $E, U \in \mathbb{R}^{|\mathcal{V}|\times d}$ with a series of $L$ transformer layers in-between \citep{attentionIsAllYouNeed}.
Each layer $l$ consists of attention heads and a multilayer perception (MLP) layer.

Given an input sequence ${\mathbf{w} = \langle w_0, ..., w_t \rangle}$, the model first applies $E$ to create an embedding ${\mathbf{x}_i \in\mathbb{R}^d}$ for each token ${w_i \in \mathbf{w}}$.
We call ${\mathbf{x}_i}$ the residual stream.

The residual stream is then updated by attention heads and MLP blocks from subsequent layers (bias terms omitted): 

$$
    \mathbf{x^{\ell+1}_i} = x^\ell_i + \texttt{MLP}^\ell(x^\ell_i + \texttt{Att}^\ell(x^\ell_i))
$$

When needed, we specify the intermittent residual stream at layer $\ell$ (after the attention head, before the MLP) as $\mathbf{x}^{\ell\_mid}$.
Per \citet{geva-etal-2022-transformer}, the updates to the residual stream from each MLP block can be further decomposed.
Namely, MLP blocks consist of two linear transformations, with point-wise activations $\sigma$ in-between:
\begin{align}
\label{eq:mlp}
    \texttt{MLP}^\ell(\mathbf{x}^\ell) = \sigma\left(W_K^\ell\mathbf{x}^{\ell} \right) W_V^\ell,
\end{align}

where $W_K^\ell, W_V^\ell \in \mathbb{R}^{d_{mlp} \times d}$.
We notate the $i$-th row in $W_K$ as $\text{MLP.}\mathbf{k}_i^{\ell}$ and refer to them as key-vectors, and the $i$-th column in $W_V$, $\text{MLP.}\mathbf{v}_i^{\ell}$, as value-vectors (we sometimes omit ``$\text{MLP}$'' and just use $\mathbf{k}_i^{\ell}, \mathbf{v}_i^\ell$).

Equation~\eqref{eq:mlp} indicates that \emph{the output of MLP blocks is the sum of its value vectors $\mathbf{v}_i$, each scaled by a coefficient value $m_i^\ell$}, where $\mathbf{m}^\ell := \sigma\left(W_K^\ell \mathbf{x}^{\ell} \right) \in \mathbb{R}^{d_{mlp}}$:

\begin{align}
\label{eq:mlp_as_sum_of_value_vectors}
  \texttt{MLP}^\ell(\mathbf{x}^{\ell}) = \sum_{i=1}^{d_{mlp}} \sigma(\mathbf{x}^{\ell} \cdot \mathbf{k}_i^{\ell}) \mathbf{v}_i^{\ell} = \sum_{i=1}^{d_{mlp}} m_i^{\ell} \mathbf{v}_i^{\ell}.
\end{align}

Put differently, the MLP block writes to the residual stream $d_{mlp}$ times, once for each value vector.
We call each of these updates a \emph{sub-update}.

All of our experiments are conducted with GPT2-medium, which has $L = 24$, $d = 1024$, and $d_{mlp} = 4096$.

\paragraph{Interpreting Value Vectors in Vocabulary Space.}
\citet{geva-etal-2022-transformer} demonstrate that for each sub-update, each value vector $\mathbf{v}_i$ either promotes or suppresses the likelihood of a token $w$ from being generated:

\begin{align}
\label{eq:value_vecs_promote}
  p&\big(w \;|\; \mathbf{x}^{\ell} + m_i^{\ell}\mathbf{v}_i^{\ell}, E\big) &\propto \exp{\big(\mathbf{e}_w\cdot\mathbf{x}^{\ell}\big)} \cdot \exp{\big(\mathbf{e}_w \cdot m_i^{\ell}\mathbf{v}_i^{\ell}\big)}
  \nonumber
\end{align}

where $\mathbf{e}_w$ is the embedding of $w$.
This indicates that when ${\mathbf{e}_w \cdot m_i^{\ell}\mathbf{v}_i^{\ell} > 0}$, the likelihood of $w$ increases, while ${\mathbf{e}_w \cdot m_i^{\ell}\mathbf{v}_i^{\ell} < 0}$ decreases the likelihood.\footnote{See Appendix for derivation.}

Further note that this dot product can be further decomposed.
Namely, $\mathbf{e}_w \cdot \mathbf{v}_i^\ell$ is a ``static'' value that does not depend on the input: only when $\mathbf{v}_i^\ell$ is scaled by $m_i$ (which is determined by the its corresponding key vector, $\mathbf{k}_i^\ell$, and the residual stream $\textbf{x}$) do we see the impact of the input towards the likelihood of $w$.

Thus the projection ${\mathbf{r}_i^\ell = E \mathbf{v}_i^\ell \in \mathbb{R}^{|\mathcal{V}|}}$ induces a ranking of tokens that get promoted by value vector $\mathbf{v}_i$, in which tokens with the highest dot products $\mathbf{e}_w \cdot \mathbf{v}_i$ are promoted most by value vector $\mathbf{v}_i$.
In Section~\ref{sec:gpt2_toxic_representation} we show value vectors that promote toxicity by applying these projections.
\section{Toxicity in Pre-trained Language Models}
\label{sec:gpt2_toxic_representation}

In this section we demonstrate how toxicity is represented and elicited in GPT2, by introducing a series of vectors that can be extracted from the language model.

\subsection{Extracting Toxic Vectors}
\label{subsec:extracting_toxic_vectors}

\paragraph{Toxicity Probe Vector.}
We start by first training a linear probe model on a binary toxicity classification task.
Namely, we use the Jigsaw toxic comment classification dataset \citep{jigsaw-toxic-comment-classification-challenge}, which consists of 561,808 comments, each of which is labeled as toxic or non-toxic.
We use a 90:10 split for training and validation.
We train our probe model, \toxicprobesymbol{}, on the residual stream in the last layer, averaged across all timesteps ($\mathbf{\bar{x}}^{L-1}$):

\begin{align}
P(\text{Toxic} | \mathbf{\bar{x}}^{L-1}) = \text{softmax}(W_{\text{Toxic}}\mathbf{\bar{x}}^{L-1}), W_{\text{Toxic}} \in \mathbb{R}^{d}
\nonumber
\end{align}

Our probe vector achieves an accuracy of 94\% on the validation split.
We view our toxic probe vector \toxicprobesymbol{} as an aggregate of all the relevant signals in the language model to classify an input as toxic.

\paragraph{Toxic Vectors in MLP Blocks.}
Given our probe vector \toxicprobesymbol{}, we can use it to find weights within the language model that promote toxicity.
Namely, \citet{geva-etal-2022-transformer} demonstrate that value vectors promote tokens at a concept-level. 
Given this, we search for value vectors that promote toxicity, by checking for all value vectors with the highest cosine similarity with \toxicprobesymbol{}.
We find that indeed, there are value vectors that promote toxic tokens (See Section~\ref{subsec:toxic_vecs_in_vocab_space}).
We notate our set of toxic value vectors as \toxicmlpsymbol{} and their corresponding key vectors as \toxicmlpkeysymbol{}.

We provide two perspectives of our \toxicmlpsymbol{} vectors: 1) when triggered, they promote the likelihood of toxic tokens to be generated, and 2) \toxicmlpsymbol{} are vectors within the model that contribute towards the \toxicprobesymbol{} direction.

\paragraph{SVD: Decomposed Toxic Vectors.}
After extracting a set of N (=128)\footnote{We experiment with different values for N, and get similar results.} \toxicmlpsymbol{} vectors, we stack them into a $N \times d$ matrix.
We then apply singular value decomposition to get decomposed singular value vectors \toxicsvdsymbol{}.
We refer to the $i$-th singular value vector as \toxicsvdsymbol{}[$i$].
We view \toxicsvdsymbol{} as basis vectors that span the toxicity representation space within the language model.

\begin{table}[t]
\caption{Top toxic vectors projected onto the vocabulary space. \textcolor{red}{WARNING: THESE EXAMPLES ARE HIGHLY OFFENSIVE.}
We note that \toxicsvdsymbol{}[2] has a particularly gendered nature.
This arises from the dataset and language model we use.}
\label{tab:projected_vocabs}
\vskip 0.15in
\begin{center}
\begin{small}
\begin{tabular}{ll}
\toprule
\textsc{Vector} & \textsc{Top tokens} \\
\midrule
\toxicprobesymbol{}         & c*nt, f*ck, a**hole, d*ck, wh*re, holes           \\
$\text{MLP.}\mathbf{v}_{770}^{19}$  & sh*t, a**, cr*p, f*ck, c*nt, garbage, trash       \\
$\text{MLP.}\mathbf{v}_{771}^{12}$  & delusional, hypocritical, arrogant, nonsense      \\
$\text{MLP.}\mathbf{v}_{2669}^{18}$  & degener, whining, idiots, stupid, smug            \\
$\text{MLP.}\mathbf{v}_{668}^{13}$  & losers, filthy, disgr, gad, feces, apes, thous    \\
$\text{MLP.}\mathbf{v}_{255}^{16}$  & disgrace, shameful, coward, unacceptable          \\
$\text{MLP.}\mathbf{v}_{882}^{12}$  & f*ck, sh*t, piss, hilar, stupidity, poop          \\
$\text{MLP.}\mathbf{v}_{1438}^{19}$  & c*m, c*ck, orgasm, missionary, anal               \\
\toxicsvdsymbol{}[0]        & a**, losers, d*ck, s*ck, balls, jack, sh*t        \\
\toxicsvdsymbol{}[1]        & sexually, intercourse, missive, rogens, nude      \\
\toxicsvdsymbol{}[2]        & sex, breasts, girlfriends, vagina, boobs          \\
\bottomrule
\end{tabular}
\end{small}
\end{center}
\vskip -0.1in
\end{table}

\begin{table}[t]
\caption{Toxicity, perplexity (PPL), and F1 after interventions or DPO.
We scale our toxic vectors such that the resulting perplexity is comparable to that of GPT2 (No Op).
$\dagger$: Not an intervention.}
\label{tab:intervene_results}
\vskip 0.15in
\begin{center}
\begin{small}
\begin{sc}
\setlength\tabcolsep{5pt}
\begin{tabular}{llccc}
\toprule
Method & Vector & Toxic & PPL & F1 \\
\midrule
No Op       & N/A                                   & 0.453 & 21.7  & 0.193 \\
\midrule

Subtract    & \toxicprobesymbol{}    & 0.245 & 23.56 & 0.193 \\
Subtract    & $\text{MLP.}\mathbf{v}_{770}^{19}$    & 0.305 & 23.30 & 0.192 \\
Subtract    & \toxicsvdsymbol{}[0]                  & 0.268 & 23.48    & 0.193 \\

\midrule

$\text{DPO}^{\dagger}$  & N/A                       &   0.208   &   23.34  & 0.195  \\
\bottomrule
\end{tabular}
\end{sc}
\end{small}
\end{center}
\vskip -0.1in
\end{table}

\begin{table*}[t]
\caption{Examples of top-k and continuations to prompts that originally elicit ``sh*t'' from GPT2, our interventions, and \dpo{}.
}
\label{tab:generation_examples}
\vskip 0.15in
\begin{center}
\begin{small}
\begin{sc}
\begin{tabular}{llll}
\toprule
Prompts & Model & Top-k & Continuations \\
\midrule
\multirow{3}{*}{\makecell[l]{\textup{``So I'm starting}\\\textup{to think she's}\\\textup{full of''}}}
    & GPT2
    & \textup{sh*t, it, crap, herself, something}
    & \textup{``sh*t. I'm not saying she's a bad person, but''} \\
    
    & GPT2 - $\text{MLP.v}_{770}^{19}$ 
    & \textup{it, herself, the, sh*t, something}
    & \textup{``it. I'm not sure if she's just being a little too''} \\

    & $\text{GPT2}_{\text{DPO}}$   
    & \textup{it, herself, sh*t, surprises, crap}
    & \textup{``it. I think she's very much aware of the fact that''} \\
    
\midrule
\multirow{3}{*}{\makecell[l]{\textup{``You are such}\\ \textup{a piece of''}}}                   
    & GPT2
    & \textup{sh*t, work, trash, garbage, a**}
    & \textup{``sh*t. You are such a piece of sh*t.''} \\

    & GPT2 - $\text{MLP.v}_{770}^{19}$ 
    & \textup{work, sh*t, ``s'', art, human}
    & \textup{``work. I've been working on this for a while''} \\
    
    & $\text{GPT2}_{\text{DPO}}$   
    & \textup{work, art, sh*t, trash, cake} 
    & \textup{``work. I have been a fan of your work for a long''} \\            
    
\midrule
\multirow{3}{*}{\textup{``Holy''}}
    & GPT2
    & \textup{sh*t, crap, cow, hell, f*ck}
    & \textup{``sh*t, I'm so glad I got this book.''} \\

    & GPT2 - $\text{MLP.v}_{770}^{19}$ 
    & \textup{Cow, Grail, cow, ``!'', Cross}
    & \textup{``Cow! I've been waiting for this for a while.''} \\
    
    & $\text{GPT2}_{\text{DPO}}$   
    & \textup{cow, crap, Grail, sh*t, smokes}
    & \textup{``cow, this is a great book! I've been reading''} \\
\bottomrule
\end{tabular}
\end{sc}
\end{small}
\end{center}
\vskip -0.1in
\end{table*}

\subsection{Toxic Vectors in Vocabulary space.}
\label{subsec:toxic_vecs_in_vocab_space}
As mentioned in Section~\ref{sec:preliminaries}, we can inspect which tokens are promoted by value vectors by projecting them onto the vocabulary space. 

Table~\ref{tab:projected_vocabs} shows the tokens with the highest dot products with our toxic vectors.
Each \toxicmlpsymbol{} and \toxicsvdsymbol{} vectors seem to encode specific dimensions of toxicity, or different contexts in which toxicity appears in pre-training data.

\subsection{Interventions Using Toxic Vectors}
\label{subsec:intervention}

To validate the role that the toxic vectors play in eliciting toxic outputs, we intervene during generation to suppress toxic outputs using each vector.
Namely, we use prompts from \textsc{RealToxicityPrompts} \citep{gehman-etal-2020-realtoxicityprompts} that elicit toxic outputs from GPT2. 
During the forward pass of the model, we intervene by simply subtracting one of the toxic vectors from the last layer:
$$
    \mathbf{x}^{L - 1} = \mathbf{x}^{L - 1} - \alpha*W,
$$

where $a$ is a heuristic scale value and $W$ is one of our toxicity vectors.

%Note that during a forward pass, value vectors $\mathbf{v}_i$ are scaled by a non-linear coefficient $\mathbf{m}_i$.
%Given that we know the location of each $MLP_{toxic}$ vector, we also experiment with turning these vectors off, by setting the corresponding coefficients $\mathbf{m}_i$ to zero.

To measure the efficacy of our interventions, we measure three metrics: toxicity, perplexity, and F1.

\paragraph{Toxicity.}
To measure toxicity, we prompt GPT2 with 
the ``challenge'' subset of \textsc{RealToxicityPrompts}, which consists of 1,199 prompts that elicit extremely toxic outputs from language models.
We follow prior work \citep{geva-etal-2022-transformer} and use Perspective API,\footnote{https://github.com/conversationai/perspectiveapi} an automated tool for toxicity detection, to assign toxicity scores to each generation.

\paragraph{Perplexity.}

To ensure that our interventions do not degrade generation quality, we also follow prior work \citep{geva-etal-2022-transformer} and measure perplexity on the Wikitext-2 dataset \citep{merity2016pointer}.

\paragraph{F1.}

In addition to perplexity, we also follow prior work \citep{dinan2020second, adolphs-etal-2023-cringe} and measure F1.
Namely, using 2,000 Wikipedia sentences as prompts, we measure the harmonic mean between precision and recall of our model's output, where precision is the fraction of generated tokens contained in the original Wikipedia continuation, and recall is the fraction of tokens in the Wikipedia continuation contained in the model's generation.

With perplexity and F1, we hope to see minimal changes after our interventions to ensure we do not affect the quality of our generations.
Table~\ref{tab:intervene_results} demonstrates the results from our interventions, while Table~\ref{tab:generation_examples} demonstrates examples of generations before and after our interventions.

Note that our interventions depend on how much we scale each vector ($\alpha$).
We choose a scalar value such that the resulting perplexity is similar to that of our post-DPO model.
For details regarding our post-DPO model see Section~\ref{sec:dpo}.

We find that subtracting toxic components from the residual stream reduces toxicity.

\section{Toxicity Alignment Using DPO}
\label{sec:dpo}

We next describe our alignment procedure using DPO.

\subsection{Background: DPO}
\label{subsec:dpo_background}

DPO relies on pairwise preference data, in which given a prompt, we have a preferred (positive) continuation and a non-preferred (negative) continuation.
Given each preference pair, the algorithm promotes the likelihood of the positive sample, while suppressing the likelihood of the negative sample, using the following loss term:

\begin{equation}
\begin{aligned}
    &\mathcal{L}_\text{DPO} = -\mathbb{E}\left[\log \sigma \left(\beta \log P - \beta \log N\right)\right] \nonumber, \\
    &P = \frac{\pi_{\theta}(y_+\mid \mathbf{w})}{\pi_{ref}(y_+\mid \mathbf{w})}, N = \frac{\pi_{\theta}(y_-\mid \mathbf{w})}{\pi_{ref}(y_-\mid \mathbf{w})},
\end{aligned}
\end{equation}

where $y_+$ and $y_-$ are preferred (nontoxic) and non-preferred (toxic) continuations of $\mathbf{w}$, $\pi_{ref}$ is the frozen weights of the original language model, and $\pi_{\theta}$ is the weights of the language model being updated (See \citet{rafailov2023direct} for details).
The algorithm promotes the likelihood of $P$, while suppressing the likelihood of $N$.

\subsection{Constructing Pairwise Toxic Data}
\label{subsec:construct_dpo_data}

We build our pairwise toxicity dataset using PPLM \citep{dathathri2019plug}.
PPLM is an attribute-controlled language generation technique, which attaches a simple linear attribute classification layer, $p(a|\mathbf{w})$ onto a language model to guide its generation.
During generation, PPLM uses the attribute classifier to compute the gradients that increases the likelihood of the language model's output to contain the desired attribute $a$, and shifts the activations in such direction (See \citet{dathathri2019plug} for details):\useshortskip

$$p(y\mid a) \propto p(y)p(a \mid y)$$ 

To generate pairwise preference data, we use sentences from Wikitext-2 \citep{merity2016pointer} as prompts.
For each prompt, we generate a positive sample using greedy sampling with GPT2, while using PPLM to generate negative (toxic) samples.
We use our toxic probe \toxicprobesymbol{} as our attribute classifier to guide towards toxic outputs.
We create 24,576 pairs of toxic and nontoxic continuations.\footnote{We release this data to enable further studies.}
We train until validation loss converges with a patience value of 10, which occurs after approximately 6,000 sample pairs.
Appendix~\ref{sec:hyperparams} has details for DPO and PPLM hyperparameters.

\begin{figure}
\vskip 0.2in
\begin{center}
\centerline{\includegraphics[width=\columnwidth]{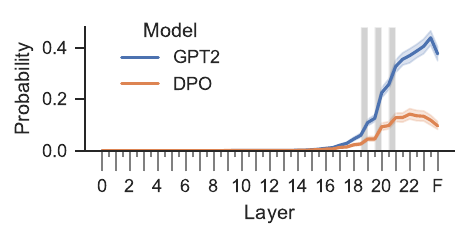}}
\caption{Logit lens on GPT2 and \dpo{}. Given 295 prompts that originally elicit ``sh*t'' as the next token, we plot the average probability of outputting ``sh*t'' from intermittent layers by applying the unembedding layer.
Minor ticks indicate $\ell\_{mid}$ layers (after attention heads, before MLP).
Shaded areas indicate layers that promote ``sh*t'' the most, which all correspond to MLP layers.
}
\label{fig:logitlens}
\end{center}
\vskip -0.2in
\end{figure}

The last row of Table~\ref{tab:intervene_results} shows the resulting toxicity, perplexity, and F1 scores of our DPO model.

Figure~\ref{fig:logitlens} shows an example of the difference in behaviors between GPT2 before and after DPO, for a specific toxic token.
Namely, we use 295 prompts from \textsc{RealToxicityPrompts} that outputs the token ``sh*t'' as the next token.
We then apply ``Logit Lens'' \citep{Nostalgebraist}, meaning we apply the unembedding layer on all intermittent layers.
This allows us to visualize the layers that promote the ``sh*t'' token.
The shared grey areas indicate the layers in which ``sh*t'' is promoted the most, which all correspond to MLP layers.
We see that post-DPO, the toxic token is promoted far less.

\section{Toxicity After DPO}
\label{sec:post_dpo}

In this section we explain how our aligned language model (\dpo{}) averts toxic outputs.

\subsection{Toxic Vectors Remain After DPO}
\label{subsec:toxic_vectors_after_dpo}

Of the toxic vectors described in Section~\ref{sec:gpt2_toxic_representation}, note that \toxicmlpsymbol{} are actual weights of the model. 
Thus we inspect how these vectors change after DPO.
%We notate the toxic MLP vectors in the original language model as \mlpgpt{} and those from the DPO model as \mlpdpo{}.

Interestingly, we find that every parameter in GPT2 and \dpo{} has barely changed, including token embeddings, MLP blocks, and attention heads.
Every parameter in GPT2 and its counterpart in \dpo{} has a cosine similarity score greater than 0.99 and on average a norm difference less than 1e-5.\footnote{The unembedding layer is the only exception, where the norm difference is less than 1e-3.}
This applies for $\text{MLP}.\mathbf{k}_{\text{Toxic}}$ and \toxicmlpsymbol{} as well -- toxic MLP vectors \textbf{do not change} from DPO.

Put differently, although toxicity is reduced by DPO, the ability to elicit toxicity with these value vectors still remain.
So how is it that \dpo{} averts toxic outputs?
Though its parameters have barely moved, below we show that their collective movement is enough to avoid toxic outputs.

\begin{figure}
\vskip 0.2in
\begin{center}
\centerline{\includegraphics[width=\columnwidth]{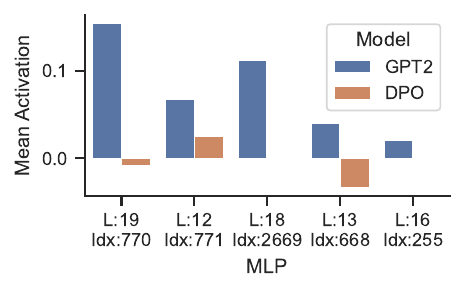}}
\caption{Mean activations for toxic vectors before and after DPO.}
\label{fig:activation_drops}
\end{center}
\vskip -0.2in
\end{figure}

\subsection{\dpo{} Avoids $\text{MLP}.\mathbf{k}_{\text{Toxic}}$ Regions}
\label{subsec:avoiding_toxic_regions}

In simplest terms, we observe a drop in activations for the toxic vectors \toxicmlpsymbol{} in \dpo{}.
Namely, using the same 1,199 prompts from \textsc{RealToxicityPrompts}, we generate 20 tokens and measure the mean activations $m_i$, or $\sigma(\mathbf{x}^\ell \cdot \text{MLP.}\mathbf{k}_i^\ell)$, of our \toxicmlpsymbol{} vectors.
Figure~\ref{fig:activation_drops} shows 5 examples of the top \toxicmlpsymbol{} vectors.

Inspired by \citet{balestriero2023characterizing}, we visualize this drop in activations with what we call ``MLP activation regions''.
An activation region of a key vector is simply a \emph{subspace} within the model's hidden space in which its vectors have high dot products to activate its corresponding value vector:

\begin{equation}
\label{eq:mlp_region}
    \gamma(\mathbf{k}_i^\ell) := \{\mathbf{g} | \mathbf{g} \in \mathbb{R}^d, \sigma(\mathbf{k}_i^\ell \cdot \mathbf{g}) > 0\},
\end{equation}
where $\sigma$ is a non-linear activation. 
Put differently, for all key-vector regions that the residual stream ``passes through'', their corresponding value-vectors are activated, scaled, and added into the residual stream.

\begin{figure}
\vskip 0.2in
\begin{center}
\centerline{\includegraphics[clip, trim=0.0cm 2.6cm 12cm 3.78cm, width=\columnwidth]{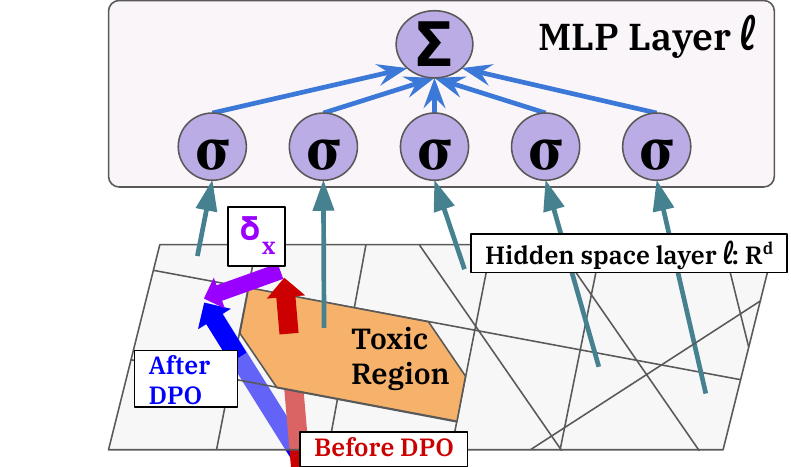}}
\caption{Visualization of residual streams before and after DPO.
We view the shift, $\delta_{\mathbf{x}}$, as an offset that allow \dpo{} to bypass regions that previously triggered toxic value vectors.
}
\label{fig:regions}
\end{center}
\vskip -0.2in
\end{figure}

We view the drop in activations as a shift in \dpo{}'s residual stream to avert the regions of toxic MLP vectors, $\gamma(\text{MLP.}\mathbf{k}_{\text{Toxic}})$.
See Figure~\ref{fig:regions}.

We formalize the shift in residual streams as following: given the residual streams at layer $\ell\_mid$ (after attention heads at layer $\ell$) for both GPT2 and \dpo{}, before $\text{MLP}_{\text{Toxic}}^{\ell}$, we notate the difference of the two residual streams as $\delta_{\mathbf{x}}^{\ell\_mid} := \mathbf{x}_{\text{DPO}}^{\ell\_mid} - \mathbf{x}_{\text{GPT2}}^{\ell\_mid}, \delta_{\mathbf{x}}^{\ell\_mid} \in \mathbb{R}^d$.
We view $\delta_{\mathbf{x}}^{\ell\_mid}$ as a vector that takes GPT2's residual stream out of the toxicity-eliciting regions, $\gamma(\text{MLP.}\mathbf{k}_{\text{Toxic}}^\ell)$.

\begin{figure}
\vskip 0.2in
\begin{center}
\centerline{\includegraphics[width=\columnwidth]{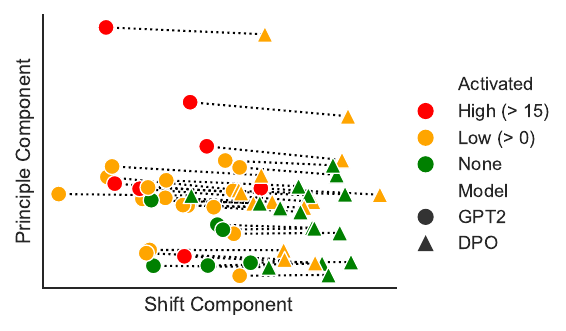}}
\caption{Linear shift of residual streams out of toxic regions.
Each point is a residual stream sampled from either $\mathbf{x}_{\text{GPT}}^{19}$ or $\mathbf{x}_{\text{DPO}}^{19}$, using \textsc{RealToxicityPrompts}, projected onto 1) $\bar{\delta}_{\mathbf{x}}^{19}$, the mean difference in residual streams, and 2) the principle component of the residual streams.
Dotted lines indicate samples from the same prompt.
Colors indicate whether each point activates $\text{MLP}_{770}^{19}$.
Note the shift from $\mathbf{x}_{\text{GPT}}^{19}$ to $\mathbf{x}_{\text{DPO}}^{19}$, but also the drop in activations.
}
\label{fig:pca}
\end{center}
\vskip -0.2in
\end{figure}

Figure~\ref{fig:pca} provides a visualization of the residual stream's shift out of toxic regions.
Namely, given prompts from \textsc{RealToxicityPrompts}, we project the residual stream from GPT2 and \dpo{} at layer 19 onto two dimensions: 1) the mean difference in the residual streams, $\bar{\delta_x^\ell}$, and the main principle component of the residual streams.\footnote{We show layer 19 because $\text{MLP.}\mathbf{v}^{19}_{770}$ is one of the most toxic vectors, but similar patterns can be found in other layers (See Appendix~\ref{sec:appx_delta_resid}).}
We further indicate whether each residual stream activates $\text{MLP.}\mathbf{v}^{19}_{770}$.
Notice both the consistent linear shift between GPT2 and \dpo{} and the drop in activations.

To understand where this shift comes from, we compute the differences in all parameter weights in GPT2 before and after DPO, and notate the differences as $\delta_{\theta}$.
We notate the difference at a specific component such as a MLP block at layer $\ell$ as $\delta_{\text{MLP}}^\ell$.

Note that as previously noted, these differences $\delta_{\theta}^\ell, \forall{\ell}$ are minimal.
Despite these minimal changes, their accumulation is sufficient in getting the residual stream out of toxic regions $\gamma(\text{MLP.}\mathbf{k}_{\text{Toxic}}^\ell)$.

Given a toxic vector \toxicmlpsymbol{} at layer $\ell$, to understand where the shift in residual stream, $\delta_{\mathbf{x}}^{\ell\_mid}$ comes from, we measure the cosine similarity between $\delta_{\mathbf{x}}^{\ell\_mid}$ and the shift in value vectors in the preceding layers, $\delta_{\text{MLP.v}}^j$:

$$
\forall{j < \ell}, \forall{i < d_{mlp}}: cos(\delta_{\mathbf{x}}^{\ell\_mid}, \delta_{\text{MLP.v}_i}^{j}).
$$

\begin{figure*}[ht]
\vskip 0.2in
\begin{center}
\centerline{\includegraphics[width=\textwidth]{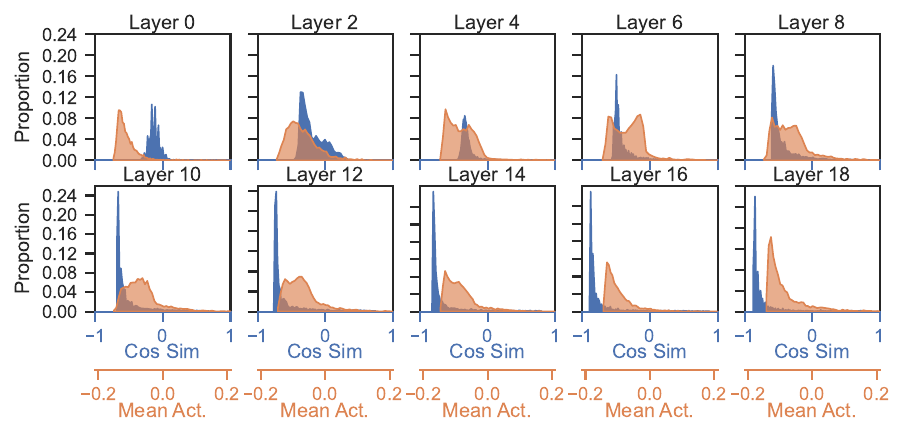}}
\caption{
The cosine similarity between $\delta_{\text{MLP}.\mathbf{v}}$ and $\delta_{\mathbf{x}}^{19}$.
Blue areas indicate the percentage of value vectors with a cosine similarity score against $\delta_{\mathbf{x}}$ as indicated by the x-axis.
Orange areas indicate the percentage of value vectors with a mean activation as indicated by the x-axis, during the forward pass of 1,199 \textsc{RealToxicityPrompts} prompts.
Value vectors shift in the opposite direction of $\delta_\mathbf{x}$, but they end up contributing towards the $\delta_\mathbf{x}$ direction because of their negative activations. 
}
\label{fig:resid_diff_plot}
\end{center}
\vskip -0.2in
\end{figure*}

To our surprise, we find that the shift in value vectors, $\delta_{MLP.v}$, have high \emph{negative} cosine similarity scores with the shift in residual streams $\delta_{\mathbf{x}}$: the value vectors in MLP blocks shift in the \emph{opposite direction} as the shift in residual stream.
The blue areas in Figure~\ref{fig:resid_diff_plot} show the cosine similarity between $\delta_{\mathbf{x}}^{19\_mid}$ and $\delta_{\text{MLP}}^{j}$.
We show layer 19 as an example because $\text{MLP.v}_{770}^{19}$ is one of the most toxic vectors, but the same pattern can be found in other layers (see Appendix~\ref{sec:appx_delta_resid_vs_delta_mlp}).
Namely, the blue areas indicate the percentage of value vectors at each layer in which their shifts have a cosine similarity score against $\delta_{\mathbf{x}}^{19\_mid}$
as indicated by the x-axis.
Note that as the layers approach layer 19, the majority of value vectors shift in the \emph{opposite} direction of $\delta_{\mathbf{x}}^{19}$.

Why the antipodal direction?
This can be explained by two facts: first, neurons in MLP blocks of language models are sparse \citep{zhang-etal-2022-moefication, li2023the}, meaning most neurons do not activate during a forward pass.
Second, the choice of the MLP's activation function $\sigma$ plays a role.
Namely, our language model uses GeLU functions \citep{hendrycks2016gaussian}.
This means that neurons that are inactive during a forward pass have a \emph{negative} value close to 0.
Thus, during the forward pass, for each value vector, the newly learned direction $\delta_{\text{MLP}.\mathbf{v}}$ gets multiplied by a very small negative scale, flips directions, and \emph{contributes} towards the $\delta_{\mathbf{x}}$ direction.
The orange areas of Figure~\ref{fig:resid_diff_plot} indicate the mean activation of each value vector, from the 1,199 prompts in \textsc{RealToxicityPrompts}.
Most of the time, value vectors have a \emph{negative} activation - thus the shift in value vectors end up \emph{contributing} towards the $\delta_{\mathbf{x}}$ direction.

To summarize, $\text{GPT2}_{\text{DPO}}$ has learned an \emph{offset}, $\delta_{\mathbf{x}}$, such that the residual stream avoids regions that promote toxicity, $\gamma(\text{MLP.}\mathbf{k}_{\text{Toxic}})$.
This learned offset is distributed across the many value vectors in earlier MLP blocks that are inactive for prompts that previously elicited toxic outputs.
By distributing this offset across numerous value vectors, the language model is able to preserve its pre-trained language modeling behavior, as individual weights are minimally affected.
However, the distributed offset allows the model to avert toxic outputs.
Note that this behavior matches precisely what the alignment objective was - to preserve as much of the pre-trained behavior, while optimizing for a reward (non-toxic outputs).

\subsection{Un-aligning \dpo{}}
\label{subsec:reactivate_toxicity}

A growing line of work finds that alignment algorithms can easily be undone or jailbroken. 
We view our findings as a mechanistic explanation for such phenomenon -- namely, in our case, the vectors that elicit toxicity are still sitting in the model, but simply not triggered.

To confirm our understanding, we demonstrate a simple way to undo alignment.
%We assume a scenario in which an adversary has access to the weights of the aligned model.
To reiterate, DPO simply learned an offset to take the residual stream $\mathbf{x}^\ell$ out of regions that trigger toxic vectors: $\gamma(\text{MLP}.\mathbf{k}_{\text{Toxic}}^\ell)$.
A simple way to re-activate toxicity is to increase those regions by scaling each key vector larger (See Equation~\ref{eq:mlp_region}).
This makes the residual streams pass through toxic regions again, thus reverting back to the pre-aligned behavior.

Table~\ref{tab:reactivate_toxicity_intervene_results} shows toxicity, perplexity, and F1 scores after scaling up as few as 7 toxic key vectors $\text{MLP}.\mathbf{k}_{\text{Toxic}}$.
We simply select 7 MLP vectors with the highest cosine similarity as our toxic probe vector, $W_{\text{Toxic}}$, and scale their key vectors by 10x.%\footnote{Note that an adversary would be able to identify these weights, by following our same procedure, i.e., by training a toxicity probe to find relevant MLP weights.}
By doing so, the model reverts back to its pre-aligned toxic behavior.
Note that increasing activation regions $\gamma$ does not have an affect on perplexity, unlike our interventions from Section~\ref{subsec:intervention}.
This is likely because the latter manipulates the residual stream directly, while scaling a key vector does not (See Equation~\ref{eq:mlp_as_sum_of_value_vectors}).

\begin{table}[t]
\caption{Un-aligning \dpo{}. By scaling toxic key vectors, and thus increasing the regions that elicit toxicity, we are able to undo the alignment learned from DPO and reactivate toxicity.}
\label{tab:reactivate_toxicity_intervene_results}
\vskip 0.15in
\begin{center}
\begin{small}
\begin{sc}
\begin{tabular}{lccc}
\toprule
Method & Toxic & PPL & F1 \\
\midrule
$\text{GPT2}_{\text{DPO}}$      & 0.208 & 23.34 & 0.195 \\
Scale $\text{MLP}.\mathbf{k}_{\text{Toxic}}$ & 0.458 & 23.30 & 0.195 \\

\midrule
$\text{GPT2}$         & 0.453 & 21.7   & 0.193 \\
\bottomrule
\end{tabular}
\end{sc}
\end{small}
\end{center}
\vskip -0.1in
\end{table}

\section{Discussion}
\label{sec:discussion}

\subsection{On Designing Robust Alignment Algorithms}
\label{subsec:discuss_jailbreak}

We view our work as providing a mechanistic explanation for why aligned models can be undone or jailbroken -- in our experiments, the regions that previously elicited toxic behavior does not change after DPO.
Rather, \dpo{} learns minimal changes spread across layers to avoid such regions and receive its reward.

With such knowledge, we conjecture that more robust alignment algorithms can be designed.
For instance, can we eliminate undesirable regions, as opposed to bypassing them?
In scenarios like ours, in which we can identify the weights that elicit undesirable outputs, what happens if we only updated those weights in isolation?
Similarly, if DPO merely learned an offset that avoids toxic regions, can we replicate this behavior by only updating the bias terms?

Alternatively, prior to deploying language models, perhaps we can add ``suppression heads'' -- layers that suppress undesirable behavior.
What would happen if we only updated late layers (or added layers) during alignment?

Lastly, can we characterize ``jailbreak-ability'' or ``unalign-ability'' of aligned models, without relying on test samples?

We leave these questions for future work.

%\subsection{Characterizing Robustness}
%\label{subsec:measuring_alignment_robustness}

%If alignment robustness is the goal, how should we characterize robustness?

%The answer may depend on the attack.
%To the best of our knowledge, current literature lacks a taxonomy of adversarial cases.
%For instance, whether an adversary only has access to an API, or to the weights of a model can lead to different attacks.
%Our safety designs may depend largely on such characteristics.

\subsection{On the Role of KL-Divergence Regularization}
\label{subsec:discuss_kl_div}

We hypothesize that the minimal changes distributed across all layers is due to the KL-divergence term that is commonly incorporated in the loss terms of RLHF algorithms.
Namely, the KL-divergence term discourages each weight from shifting too drastically, in order to preserve its capabilities learned during pre-training.

Similar to our work, \citet{jain2023mechanistically} fine-tunes a language model on synthetic tasks to study the changes in its mechanisms.
Interestingly, unlike our findings, the authors demonstrate that the model simply learns ``wrappers'' at late layers that optimize for each task.

We find this difference in model training behavior interesting, and conjecture that the KL-divergence term may play a role in this difference.
Note that fine-tuning typically does not entail a KL-divergence term.
Perhaps this allows the model to make drastic and localized changes, such as in late layers, as opposed to distributed, minimal changes.

%\subsection{On Generalizing Findings from Our Case Study}
%\label{subsec:generalizing_findings_from_case_study}

%We note that our findings are from a case study, on a specific language model, task, and data set.

%An obvious extension would be to study a larger model.
%While in principle, we do not expect the mechanisms of DPO to differ for a larger model, we have also seen that the scale of models lead to surprising results \citep{hoffmann2022training, wei2022emergent}.
%We leave this for future work.

\section{Related Work}
\label{sec:related_work}

%JKK: Better to leave this out and get straight into it
%In this section we discuss related work.

\subsection{Alignment Algorithms}
\label{subsec:related_work_alignment_algorithms}

Numerous alignment algorithms have been proposed, and the choice of algorithm may largely depend on the type of data available.
Perhaps most commonly, human feedback data is used \citep{stiennon2020learning, ouyang2022training, touvron2023llama} for methods such as PPO \citep{schulman2017proximal} or DPO \citep{rafailov2023direct}.
When labels for only undesirable behavior is available, algorithms like unlikelihood training \citep{Welleck2020Neural} or Cringe \citep{adolphs-etal-2023-cringe, xu2023some} can be used.
We study DPO because it is easy to use and currently widely used.

\subsection{Mechanistic Interpretability}
\label{subsec:related_work_mech_interp}

The goal of mechanistic interpretability is largely to reverse engineer model behaviors \citep{olah2020zoom, elhage2021mathematical, geva-etal-2021-transformer}.
By doing so, researchers have uncovered various interpretable and controllable representations, such as world models \citep{li2023emergent, nanda2023emergent}, ``truthfulness'' \citep{li2023inference}, knowledge \citep{meng2022locating, hernandez2023linearity, burns2023discovering, geva2023dissecting}, linguistic properties \citep{conneau-etal-2018-cram, tenney-etal-2019-bert}, or even tasks \citep{ilharco2022editing, hendel-etal-2023-context, todd2023function}.

Rather than probing for specific representations, researchers have also characterized the representations of language models from a geometric perspective \citep{park2023linear}.
\citet{balestriero2023characterizing} demonstrate a geometric characterization that can be used to extract feature representations that solve toxicity detection.

Similar to our work, \citet{jain2023mechanistically} study the mechanisms in which fine-tuning on synthetic tasks alters the model's capabilities.
We study the effects of RLHF on a more realistic, natural language setting.

\subsection{Jailbreaking Aligned Models}
\label{subsec:related_work_jailbreak}

Researchers demonstrated that aligned models can be surprisingly easily jailbroken \citep{wallace-etal-2019-universal, zou2023universal, wei2023jailbroken, carlini2023are}.
Such adversarial attacks typically involve searching for prompts that can elicit previously unlearned behaviors, or even personal information \citep{nasr2023scalable}.
\citet{carlini2023are} show that multimodal models can also be jailbroken.
\citet{wei2023jailbroken} provide hypotheses, backed by empirical studies, as to why language models can be jailbroken.

In a similar vein to jailbreaks, numerous researchers have demonstrated that aligned models can easily be un-aligned \citep{yang2023shadow, qi2023fine}, sometimes with as few as 100 fine-tuning examples.
We view our work as adding a mechanistic understanding of such phenomena.

\section{Conclusion}
\label{sec:conclusion}

In this work we studied the mechanisms by which alignment algorithms unlearn a capability, taking DPO and toxicity as a case study.
First, we uncovered how toxicity is represented and elicited in a pre-trained language model.
We find numerous vectors in MLP blocks that promote toxicity.
Simply subtracting these vectors from the residual stream can suppress toxic outputs.

Second, we applied DPO to our language model, using PPLM to carefully craft pairs of toxic and non-toxic continuations to Wikipedia prompts.

Third, we studied how our aligned model \dpo{} averts toxicity.
Rather than removing the regions that elicit toxicity, \dpo{} bypasses them by learning an \emph{offset}.
Such an offset is distributed amongst multiple value vectors, allowing minimal changes to every weight.
This allows the model to preserve its pre-trained behavior, while averting toxic outputs, which matches the objective of the DPO loss.

Given this understanding, we demonstrated how to break the alignment of \dpo{}, reverting it back to its toxic behavior.
Namely, we simply increase the regions that elicit toxicity, by scaling their corresponding key vectors.

We view our findings as a mechanistic case study for why aligned models can be jailbroken, and hope that this can lead to more robust alignment algorithms.
Our code, models, and data can be found at \url{https://github.com/ajyl/dpo_toxic}.

\section*{Acknowledgements}

We thank Ekdeep Singh Lubana for fruitful discussions, and Santiago Serra Castro for helping with figures.
This work was supported via NSF under grant \#2306372.
%\textbf{Do not} include acknowledgements in the initial version of
%the paper submitted for blind review.

%If a paper is accepted, the final camera-ready version can (and
%probably should) include acknowledgements. In this case, please
%place such acknowledgements in an unnumbered section at the
%end of the paper. Typically, this will include thanks to reviewers
%who gave useful comments, to colleagues who contributed to the ideas,
%and to funding agencies and corporate sponsors that provided financial
%support.

\bibliography{main}
\bibliographystyle{icml2024}

%%%%%%%%%%%%%%%%%%%%%%%%%%%%%%%%%%%%%%%%%%%%%%%%%%%%%%%%%%%%%%%%%%%%%%%%%%%%%%%
%%%%%%%%%%%%%%%%%%%%%%%%%%%%%%%%%%%%%%%%%%%%%%%%%%%%%%%%%%%%%%%%%%%%%%%%%%%%%%%
% APPENDIX
%%%%%%%%%%%%%%%%%%%%%%%%%%%%%%%%%%%%%%%%%%%%%%%%%%%%%%%%%%%%%%%%%%%%%%%%%%%%%%%
%%%%%%%%%%%%%%%%%%%%%%%%%%%%%%%%%%%%%%%%%%%%%%%%%%%%%%%%%%%%%%%%%%%%%%%%%%%%%%%
\newpage
\appendix
\onecolumn
\section{Projecting Value Vectors onto Vocabulary Space}
\label{sec:appx_vocab_projection}

In this section we provide details from \citet{geva-etal-2022-transformer} that demonstrate that MLP value vectors promote or suppress the likelihood of tokens.

We start from Equation~\ref{eq:mlp_as_sum_of_value_vectors}:

$$
  \texttt{MLP}^\ell(\mathbf{x}^{\ell}) = \sum_{i=1}^{d_{mlp}} \sigma(\mathbf{x}^{\ell} \cdot \mathbf{k}_i^{\ell}) \mathbf{v}_i^{\ell} = \sum_{i=1}^{d_{mlp}} m_i^{\ell} \mathbf{v}_i^{\ell}.
$$

Thus, we can consider the update from $\texttt{MLP}^\ell$ as $d_{mlp}$ \emph{sub-updates}, each sub-update being $m^\ell_i\mathbf{v}^\ell_i$.

We can then analyze the influence that each sub-update has on the output distribution, or the probability of generating token $w \in V$ (taken from \citet{geva-etal-2022-transformer}):

\begin{align}
    p&\big(w \;|\; \mathbf{x}^{\ell} + m_i^{\ell}\mathbf{v}_i^{\ell}, E\big) = \frac{\exp{\big(\mathbf{e}_w\cdot\mathbf{x}^{\ell}+\mathbf{e}_w\cdot m_i^{\ell}\mathbf{v}_i^{\ell}\big)}}{Z\big(E (\mathbf{x}^{\ell} + m_i^{\ell}\mathbf{v}_i^{\ell}))} \propto \exp{\big(\mathbf{e}_w\cdot\mathbf{x}^{\ell}\big)} \cdot \exp{\big(\mathbf{e}_w \cdot m_i^{\ell}\mathbf{v}_i^{\ell}\big)}
\end{align}

where $\mathbf{e}_w$ is the token embedding of $w$, and $Z$ is the softmax normalization factor.
This indicates that when ${\mathbf{e}_w \cdot m_i^{\ell}\mathbf{v}_i^{\ell} > 0}$, the likelihood of $w$ increases, while ${\mathbf{e}_w \cdot m_i^{\ell}\mathbf{v}_i^{\ell} < 0}$ decreases the likelihood.

\section{Shift in Residual Streams}
\label{sec:appx_delta_resid}

\begin{figure*}[ht]
\vskip 0.2in
\begin{center}
\centerline{\includegraphics[width=\columnwidth]{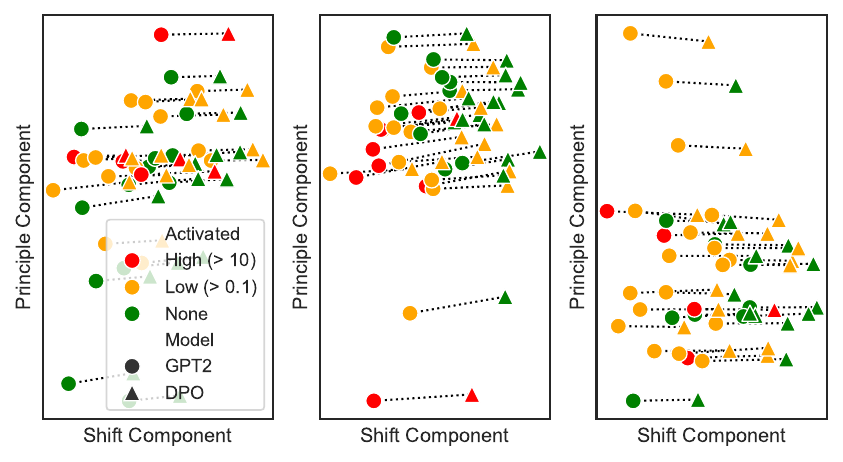}}
\caption{
Shift in residual streams at layer 12, 18, and 13 (we show these three layers because $\text{MLP}.\mathbf{v}^{12}_{771}$, $\text{MLP}.\mathbf{v}^{18}_{2669}$, and $\text{MLP}.\mathbf{v}^{13}_{668}$ are the next three vectors with highest cosine similarity with $W_{\text{Toxic}}$. See Table~\ref{tab:projected_vocabs}, Figure~\ref{fig:activation_drops}.
}
\label{fig:appx_pca}
\end{center}
\vskip -0.2in
\end{figure*}

In this section we provide more examples of residual streams shifting out of toxic regions.
See Figure~\ref{fig:appx_pca}

\section{Shifts in Residual Streams vs. Shifts in MLP Value Vectors.}
\label{sec:appx_delta_resid_vs_delta_mlp}

In this section we provide more examples of how MLP value vectors contribute in the $\delta_{\mathbf{x}}$ direction at different layers.

\begin{figure*}[ht]
\vskip 0.2in
\begin{center}
\centerline{\includegraphics[width=\textwidth]{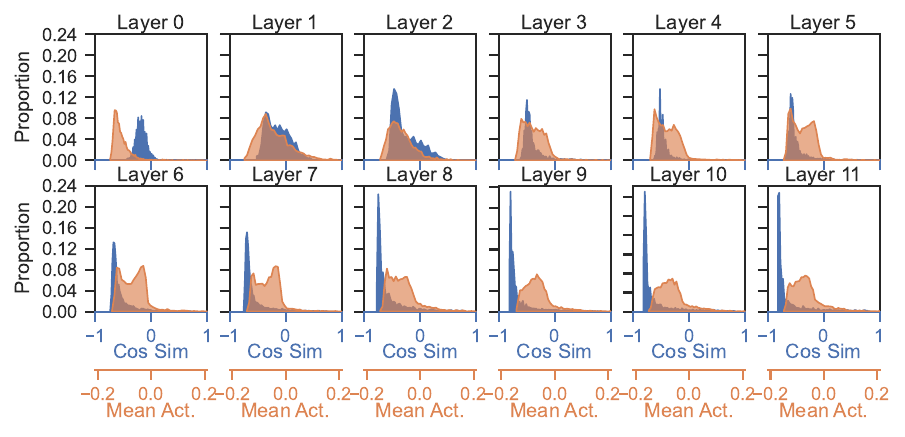}}
\caption{
Shift in residual streams at layer 12 vs. shift in MLP value vectors ($\delta_{\mathbf{x}}^{12}$ vs. $\delta_{\text{MLP}}$). 
}
\label{fig:resid_diff_layer_12}
\end{center}
\vskip -0.2in
\end{figure*}

\begin{figure*}[ht]
\vskip 0.2in
\begin{center}
\centerline{\includegraphics[width=\textwidth]{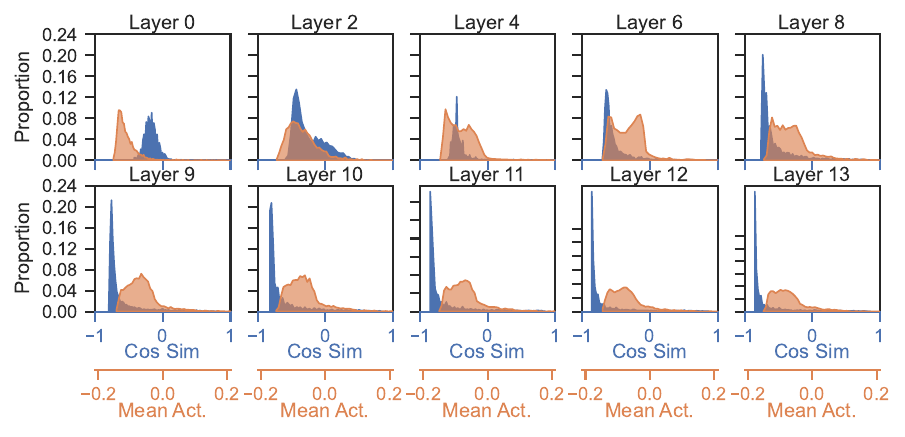}}
\caption{
Shift in residual streams at layer 14 vs. shift in MLP value vectors ($\delta_{\mathbf{x}}^{14}$ vs. $\delta_{\text{MLP}}$). 
}
\label{fig:resid_diff_layer_14}
\end{center}
\vskip -0.2in
\end{figure*}

\begin{figure*}[ht]
\vskip 0.2in
\begin{center}
\centerline{\includegraphics[width=\textwidth]{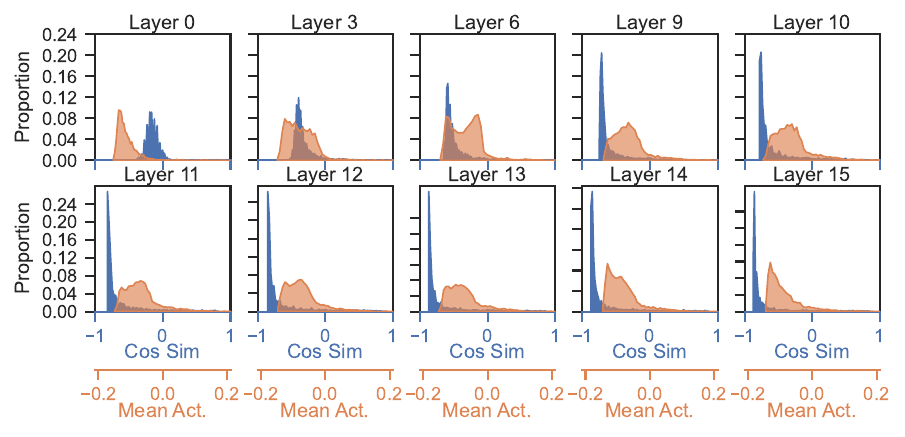}}
\caption{
Shift in residual streams at layer 16 vs. shift in MLP value vectors ($\delta_{\mathbf{x}}^{16}$ vs. $\delta_{\text{MLP}}$). 
}
\label{fig:resid_diff_layer_16}
\end{center}
\vskip -0.2in
\end{figure*}

\begin{figure*}[ht]
\vskip 0.2in
\begin{center}
\centerline{\includegraphics[width=\textwidth]{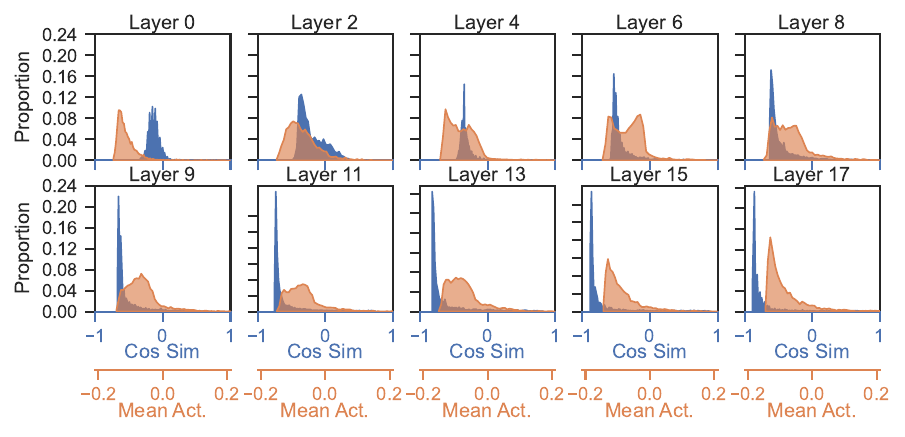}}
\caption{
Shift in residual streams at layer 18 vs. shift in MLP value vectors ($\delta_{\mathbf{x}}^{18}$ vs. $\delta_{\text{MLP}}$). 
}
\label{fig:resid_diff_layer_18}
\end{center}
\vskip -0.2in
\end{figure*}

\section{Hyperparameters}
\label{sec:hyperparams}

Tables \ref{tab:hyperparams_dpo}, and \ref{tab:hyperparams_pplm} contain the hyperparameters used for our toxic probe, DPO, and PPLM, respectively.

\begin{table}[t]
\caption{Hyperparameters: DPO.}
\label{tab:hyperparams_dpo}
\vskip 0.15in
\begin{center}
\begin{small}
\begin{sc}
\begin{tabular}{ll}
\toprule
Hyperparameter & Value \\
\midrule
Learning Rate   & 1e-6      \\
Batch Size      & 4         \\
Optimizer       & RMSprop   \\
Gradient accumulation steps &   1           \\
Max gradient Norm           &   10          \\
Validation Metric           &   loss/valid  \\
Validation Patience         &   10          \\
DPO Beta                    &   0.1         \\
\bottomrule
\end{tabular}
\end{sc}
\end{small}
\end{center}
\vskip -0.1in
\end{table}

\begin{table}[t]
\caption{Hyperparameters: PPLM.}
\label{tab:hyperparams_pplm}
\vskip 0.15in
\begin{center}
\begin{small}
\begin{sc}
\begin{tabular}{ll}
\toprule
Hyperparameter & Value \\
\midrule
Step Size       &   0.4 \\
Temperature     &   1   \\
Top K           &   10  \\
Num Iterations  &   50  \\
Window Length   &   0   \\
Horizon Length  &   1   \\
Decay           &   False   \\
Gamma           &   1   \\
GM Scale        &   0.95    \\
KL Scale        &   0.1 \\
\bottomrule
\end{tabular}
\end{sc}
\end{small}
\end{center}
\vskip -0.1in
\end{table}

%%%%%%%%%%%%%%%%%%%%%%%%%%%%%%%%%%%%%%%%%%%%%%%%%%%%%%%%%%%%%%%%%%%%%%%%%%%%%%%
%%%%%%%%%%%%%%%%%%%%%%%%%%%%%%%%%%%%%%%%%%%%%%%%%%%%%%%%%%%%%%%%%%%%%%%%%%%%%%%

\end{document}